\documentclass[conference]{IEEEtran}

\usepackage{graphicx}
\usepackage{amsmath}
\usepackage{algorithmic}

\begin{document}

\title{Adaptive Gaussian Fuzzy Classifier for Real-Time Emotion Recognition in Computer Games}

\author{\IEEEauthorblockN{Daniel Leite, Volnei Frigeri Jr., and Rodrigo Medeiros}
\IEEEauthorblockA{Department of Automatics, Federal University of Lavras (UFLA), Brazil \\ Email: daniel.leite@ufla.br, vjunior.frigeri@gmail.com, rodrigo.nmedeiros19@gmail.com}}

\maketitle

\begin{abstract}
Human emotion recognition has become a need for more realistic and interactive machines and computer systems. The greatest challenge is the availability of high-performance algorithms to effectively manage individual differences and nonstationarities in physiological data streams, i.e., algorithms that self-customize to a user with no subject-specific calibration data. We describe an evolving Gaussian Fuzzy Classifier (eGFC), which is supported by an online semi-supervised learning algorithm to recognize emotion patterns from electroencephalogram (EEG) data streams. We extract features from the Fourier spectrum of EEG data. The data are provided by 28 individuals playing the games `Train Sim World', `Unravel', `Slender The Arrival', and `Goat Simulator' -- a public dataset. Different emotions prevail, namely, boredom, calmness, horror and joy. We analyze the effect of individual electrodes, time window lengths, and frequency bands on the accuracy of user-independent eGFCs. We conclude that both brain hemispheres may assist classification, especially electrodes on the frontal (Af3-Af4), occipital (O1-O2), and temporal (T7-T8) areas. We observe that patterns may be eventually found in any frequency band; however, the Alpha (8-13Hz), Delta (1-4Hz), and Theta (4-8Hz) bands, in this order, are the highest correlated with emotion classes. eGFC has shown to be effective for real-time learning of EEG data. It reaches a 72.2\% accuracy using a variable rule base, 10-second windows, and 1.8ms/sample processing time in a highly-stochastic time-varying 4-class classification problem.
\end{abstract}

\IEEEpeerreviewmaketitle

\section{Introduction}

Human emotions can be noticed by machines through non-physiological data, such as facial expressions, gestures, body language, eye blink count, voice tones; and physiological data by using EEG, electromyogram (EMG), and electrocardiogram (ECG) \cite{Song:18}\cite{VaMeMi:20}. Computer vision and audition, physiological features, and brain-computer interface (BCI) are typical approaches to emotion recognition. The potential outcomes of non-physiological and physiological data analysis spread across a variety of domains in which machine intelligence has offered decision-making support; control of mechatronics; and realism, efficiency, and interaction \cite{Alarcao}.

Classifiers of EEG data are usually based on Support Vector Machines (SVM) using different kernels, the $k$-Nearest Neighbors (kNN) algorithm, Linear Discriminant Analysis (LDA), Naive Bayes (NB), and the Multi-Layer Perceptron (MLP) neural network \cite{Alarcao} \cite{Gu}. Deep learning has also been applied to affective computing from physiological data. A deep neural network to classify the states of relaxation, anxiety, excitement, and fun by means of skin conductance and pulse signals, with similar accuracy to those of shallow methods, is given in \cite{Martinez}. A Deep Belief Network (DBN), i.e., a probabilistic generative deep model, to classify positive, neutral, and negative emotions is proposed in \cite{Zheng2}. Selection of electrodes and frequency bands is performed through the weight distributions obtained from the trained DBN, being differential asymmetries between the left and right brain hemispheres relevant features. A Dynamical-Graph Convolutional Neural Network (DG-CNN), trained by error back-propagation, learns an adjacency matrix among EEG channels to outperform DBN, Transductive SVM, Transfer Component Analysis (TCA), and other methods on benchmark EEG datasets in \cite{Song:18}.

Fuzzy methods to handle uncertainties in emotion recognition, specially using speech data and facial images have been proposed. A Two-Stage Fuzzy Fusion strategy combined with a CNN (TSFF-CNN) is described in \cite{WuSu:20}. Facial expressions and speech modalities are aggregated for a final classification decision. The method manages ambiguity and uncertainty of emotional state information. The TSFF-CNN outperformed non-fuzzy deep methods. In \cite{Zhang:16}, a Fuzzy multi-class SVM method uses features from the Biorthogonal wavelet transform applied over facial images to detect emotions, namely, happiness, sadness, surprisingness, angriness, disgustingness, and fearfulness. An Adaptive Neuro-Fuzzy Inference System (ANFIS) that combines facial expression and EEG features have shown to be superior to single-source-based classifiers in \cite{Lee:14}. The ANFIS model identifies the valence status stimulated by watching movies.

Multi-scale Inherent Fuzzy Entropy (I-FuzzyEn) is a method that uses empirical mode decomposition and fuzzy membership functions to evaluate the complexity of EEG data \cite{ZCao}. Complexity estimates are useful as a health bio-signature. FuzzyEn has shown to be more robust than analogous non-fuzzy methods. The Online weighted Adaptation Regularization for Regression (OwARR) algorithm \cite{DWU} aims to estimate driver drowsiness from EEG data. OwARR uses fuzzy sets to realize part of a regularization procedure. Some offline training steps are needed to select domains. The idea is to reduce subject-specific calibration from EEG data. An ensemble of models using swarm-optimized fuzzy integral for motor imagery recognition and robotic arm control is given in \cite{SLWu}. The ensemble uses Sugeno or Choquet fuzzy integral, supported by particle swarm optimization (PSO); it identifies mental representation of movements.

These classifier-design methods indirectly assume stationary data sources since models are expected to keep their training performance during online operation using fixed structure and parameters. However, physiological data change due to movement artifacts, electrode and channel configuration, and environmental conditions. Fatigue, attention and stress also affect both user-dependent and user-independent classifiers in an uncertain way. In the present study we cope with real-time emotion classification from EEG data streams (visual and auditory evoked potentials) generated by computer game players. We summarize a recently-proposed semi-supervised evolving Gaussian Fuzzy Classification (eGFC) method \cite{FUZZlog2}. eGFC is an instance of Evolving Granular System \cite{Lei:12} \cite{SkIgSaLeLuGo:19}, which is a general-purpose online learning framework, i.e., a family of algorithms and  methods to autonomously construct classifiers, regressors, predictors, and controllers in which any aspect of a problem may assume a non-pointwise (e.g., interval, fuzzy, rough, statistical) uncertain characterization, including data, parameters, features, learning equations, and covering regions \cite{Garcia} \cite{LeAnSkGo:20} \cite{LeCoGo:13}. eGFC has been applied to power quality classification in smart grids \cite{FUZZlog2}; and anomaly detection in data centers \cite{FUZZlog}. 

We use a public EEG dataset \cite{AlGoTu:20}, which contains ordered samples obtained from individuals exposed to visual and auditory stimuli. Brain activity is recorded by the 14 channels of the \textit{Emotiv EPOC+} EEG device. We pre-process raw data by means of time windows and frequency filters. We extract ten features per EEG channel, namely, the maximum and mean values from the Delta (1-4Hz), Theta (4-8Hz), Alpha (8-13Hz), Beta (13-30Hz), and Gamma (30-64Hz) bands -- 140 features in total. A unique user-independent eGFC model is developed from scratch based on sequential data from 28 consecutive players. After recognizing a predominant emotion according to the Arousal-Valence system, i.e., boredom, calmness, horror, or joy, which are related to specific computer games, then a feedback can be implemented in a real or virtual environment to promote a higher level of realism and interactivity. The feedback step is out of the scope of this study.

Our contribution in relation to the literature are: (i) a new user-independent time-varying fuzzy classifier, eGFC, supplied with online learning to deal with uncertainties and nonstationarities of EEG data. eGFC incorporates new spatio-temporal patterns with no human intervention by using adaptive granularity and interpretable rule base. Storing data samples, and \textit{a priori} knowing the task and number of classes are needless; (ii) an analysis of the effect of time window lengths, brain regions, frequency bands, and dimensionality reduction on the classification performance.

\section{Evolving Gaussian Fuzzy Classifier}
\label{sec:sluc}

We summarize the semi-supervised evolving classification method, eGFC \cite{FUZZlog2}. Although eGFC handles partially labeled data, we assume labels become available some time steps after the class estimate. A single supervised learning step is given using an input-output pair when the output is available. eGFC uses Gaussian membership functions to cover the data domain with granules. Recursive equations construct its rule base and updates granules to deal with changes  and provide nonlinear and smooth class boundaries \cite{FUZZlog2} \cite{FUZZlog}.

\subsection{Rule Structure}
\label{sec:GaussFunc}

Rules are created and updated depending on the behavior of the system over time. An eGFC rule, say $R^i$, is

\vspace{7pt}

\noindent IF $\left((x_1 \textrm{ is } A_1^i) \textrm{ AND } ... \textrm{ AND } (x_n \textrm{ is } A_n^i)\right), \textrm{THEN } (y \textrm{ is } C^i)$

\vspace{7pt}

\noindent in which $x_j$, $j = 1, ..., n$, are features, and $y$ is a class. The data stream is denoted ${(\textbf{x}, y)}^{[h]}, h = 1, ...$ Moreover, $A_j^i$,  $\forall j$; $i = 1, ..., c$, are Gaussian functions built and updated incrementally; $C^i$ is a class label. $A_j^i = G(\mu_j^i, \sigma_j^i)$ has height 1, and is characterized by the modal value $\mu_j^i$ and dispersion $\sigma_j^i$. Rules $R^i$, $\forall i$, set up a zero-order Takagi-Sugeno model. The number of rules, $c$, is variable. We give a semi-supervised way of constructing eGFC from scratch.

\subsection{Adding Classification Rules}
\label{sec:AddRule}

Rules are created and evolved as data are available. A new granule, $\gamma^{c+1}$, and rule, $R^{c+1}$, are created if none of the existing rules $\{R^1, ..., R^c\}$ are sufficiently activated by $\textbf{x}^{[h]}$. 

Let $\rho^{[h]} \in [0,1]$ be an adaptive threshold. If

\begin{equation}
T\left(A_1^i(x_1^{[h]}),...,A_n^i(x_n^{[h]})\right)\leq \rho^{[h]}, ~ \forall i, \label{activ}
\end{equation}

\noindent in which $T$ is the minimum triangular norm, then the eGFC structure is expanded. $\rho^{[h]}$ dictates how large granules can be. Different values result in different granular perspectives \cite{LeCoGo:13}. Section \ref{sec:dispersion} gives recursive equations to update $\rho^{[h]}$.

A new $\gamma^{c+1}$ has membership functions $A_j^{c + 1}$, $\forall j$, with

\begin{equation}
\mu_j^{c+1} = x_j^{[h]}, \textrm{ and } \sigma_j^{c+1} = 1/2\pi. \label{eq6}
\end{equation}

\noindent The intuition is to start big. Dispersions diminish when new samples activate the same granule. This strategy is appealing for a compact model structure \cite{LeAnSkGo:20}.

The class $C^{c+1}$ is initially undefined. If the corresponding output, $y^{[h]}$, associated to $\textbf{x}^{[h]}$, becomes available, then

\vspace{-1pt}

\begin{equation}
C^{c+1} = y^{[h]}. \label{eq8}
\end{equation}

\noindent Otherwise, the first labeled sample that arrives after the $h$-th time step, and activates the rule $R^{c+1}$ according to \eqref{activ}, is used to define its class.

In case a labeled sample activates a labeled rule, but their labels are different, then a new (partially overlapped) granule is created to represent new information. Partially overlapped Gaussian granules, tagged with different labels, tend to have their modal values withdrawn and dispersions reduced by the updating procedure (Section \ref{sec:ParAdapt}). With this initial rule parameterization, preference is given to the design of granules balanced along its dimensions. eGFC realizes the principle of justifiable granularity \cite{Wang}, but allows Gaussians to find more appropriate places and dispersions.

\subsection{Updating Parameters}
\label{sec:ParAdapt}

The $i$-th rule is candidate to be updated if it is sufficiently activated by an unlabeled sample, $\textbf{x}^{[h]}$, according to

\begin{equation}
min\left(A_1^i(x_1^{[h]}),...,A_n^i(x_n^{[h]})\right) > \rho^{[h]}. \label{eq9}
\end{equation} 

\noindent Only the most active rule, $R^{i^*}$, is chosen for adaptation in case two or more rules reach the $\rho^{[h]}$ level for the unlabeled $\textbf{x}^{[h]}$. For a labeled sample, i.e., for pairs $(\textbf{x},y)^{[h]}$, the class of the most active rule $R^{i^*}$, if defined, must match $y^{[h]}$. Otherwise, the second most active rule among those that reached the $\rho^{[h]}$ level is chosen for adaptation, and so on. If none of the rules are apt, then a new one is created (Section \ref{sec:AddRule}).

To include $\textbf{x}^{[h]}$ in $R^{i^*}$, the learning algorithm updates the modal values and dispersions of $A_j^{i^*}$, $\forall j$,

\begin{equation}
\mu_j^{i^*}(new) = \frac {(\varpi^{i^*}-1) \mu_j^{i^*}(old) + x_j^{[h]}}{\varpi^{i^*}}, \label{eq10}
\end{equation} 

\noindent and

\vspace{-3pt}

\begin{eqnarray}
\sigma_j^{i^*}(new) &=& \biggl( \frac {(\varpi^{i^*}-1)}{\varpi^{i^*}} ~ \left(\sigma_j^{i^*}(old)\right)^2 ~+~ \nonumber \\ && + \frac {1}{\varpi^{i^*}} \left(x_j^{[h]} - \mu_j^{i^*}(old)\right)^2 \biggr)^{1/2}, ~~~ \label{eq11}
\end{eqnarray} 

\noindent in which $\varpi^{i^*}$ is the number of times the $i^*$-th rule was updated previously. Notice that \eqref{eq10}-\eqref{eq11} are recursive and, therefore, do not require data storage. We keep $\sigma_j^{i^*}$ between a minimum value, $1/(4\pi)$; and the Stigler limit, $1/(2\pi)$ \cite{LeAnSkGo:20}.

\subsection{Time-Varying Granularity}
\label{sec:dispersion}

Let the activation threshold, $\rho^{[h]}$, be time-varying \cite{Garcia}. The threshold assumes values in the unit interval according to the overall average dispersion

\vspace{-2pt}

\begin{equation}
\sigma_{avg}^{[h]} = \frac{1}{cn} \sum\limits_{i=1}^c \sum\limits_{j=1}^n \sigma^{i[h]}_j, \label{eq12}
\end{equation} 

\noindent in which $c$ and $n$ are the number of rules and features. Thus,

\begin{equation}
\rho(new) = \frac{\sigma_{avg}^{[h]}}{\sigma_{avg}^{[h-1]}} ~ \rho(old). \label{eq13}
\end{equation} 

Given $\textbf{x}^{[h]}$, rules' activation degrees are compared to $\rho^{[h]}$ to decide between parametric or structural change of the eGFC. We suggest $\rho^{[0]} = 0.1$ as starting value.

\subsection{Merging}
\label{sec:mergGranules}

A distance measure between Gaussian objects is

\vspace{-2pt}

\begin{equation}
d(\gamma^{i_1}, \gamma^{i_2}) = \frac {1}{n} ~\sum_{j=1}^n | \mu_j^{i_1} - \mu_j^{i_2} | + \sigma_j^{i_1} + \sigma_j^{i_2} - 2 \sqrt{\sigma_j^{i_1} \sigma_j^{i_2}} \label{eq14}
\end{equation}

\noindent eGFC may combine the pair of granules with the smallest $d(.)$. The underlying granules must be either unlabeled or tagged with the same class label. The merging decision is based on a threshold, $\Delta$, or expert judgment regarding the suitability of combining granules to have a more compact model. For data within the unit hypercube, we use $\Delta = 0.1$ as default, which means that the candidate granules should be quite similar and, in fact, carry quite similar information.

A new granule, say $\gamma^i$, which results from $\gamma^{i_1}$ and $\gamma^{i_2}$, is built by Gaussians with modal values

\begin{equation}
\mu_j^i = \frac{\frac{\sigma_j^{i_1}}{\sigma_j^{i_2}}\mu_j^{i_1} + \frac{\sigma_j^{i_2}}{\sigma_j^{i_1}}\mu_j^{i_2}}{\frac{\sigma_j^{i_1}}{\sigma_j^{i_2}} + \frac{\sigma_j^{i_2}}{\sigma_j^{i_1}}}, ~ \forall j,  \label{eq15}
\end{equation}

\noindent and dispersion

\begin{equation}
\sigma_j^i = \sigma_j^{i_1} + \sigma_j^{i_2}, ~ \forall j. \label{eq16}
\end{equation}

\subsection{Deleting Rules}
\label{sec:delRules}

A rule is removed from the eGFC model if it is inconsistent with the current environment. In other words, if a rule is not activated for a number of time steps, say $h_r$, then it is deleted from the rule base. However, if a class is rare, then it may be the case to set $h_r = \infty$, and keep the inactive rules. Removing rules periodically helps to keep the rule base updated.

\subsection{Online Learning from Data Stream}
\label{sec:learning}

See \cite{FUZZlog2} or  \cite{FUZZlog} for a pseudocode of the semi-supervised algorithm to construct and update an eGFC on the fly.

\section{Methodology} \label{methodology}

We want to classify emotions from EEG data (visual and auditory evoked potentials). We describe how features are extracted and models are evaluated based on particular channels and multivariable BCI system.

\subsection{About the Data}

\begin{figure}[b]
	\begin{center}
		\includegraphics[width=6.1cm]{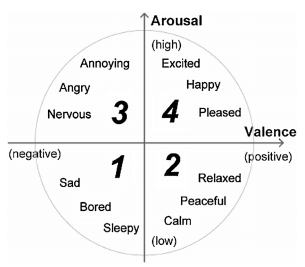}
		\caption{The Arousal-Valence model of human emotions}
		\label{Fig2}
	\end{center}
\end{figure}

The eGFC is evaluated from EEG streams generated by individuals playing computer games. A game motivates predominantly a particular emotion, according to the quadrants of the Arousal-Valence circle \cite{Russell}. On the left side of the valence dimension are negative emotions, e.g., `angry', `bored', and relative adjectives, whereas on the right side are positive emotions, e.g., `happy', `calm', and relative adjectives. The upper part of the arousal dimension characterizes excessive emotional arousal or behavioral expression, while the lower part designates apathy. We assign a number to the quadrants (Figure \ref{Fig2}) according to our classification problem. The classes are: bored, calm, anger, and happy. As the players effectively change the outcome of the game, their mental activity is high. They cognitively process images, build histories mentally, and evaluate characters and situations. At first, emotions are not prone to any quadrant  \cite{VaMeMi:20} \cite{LaFeSa:14}.

\begin{figure}[h]
	\begin{center}
		\includegraphics[width=5.4cm]{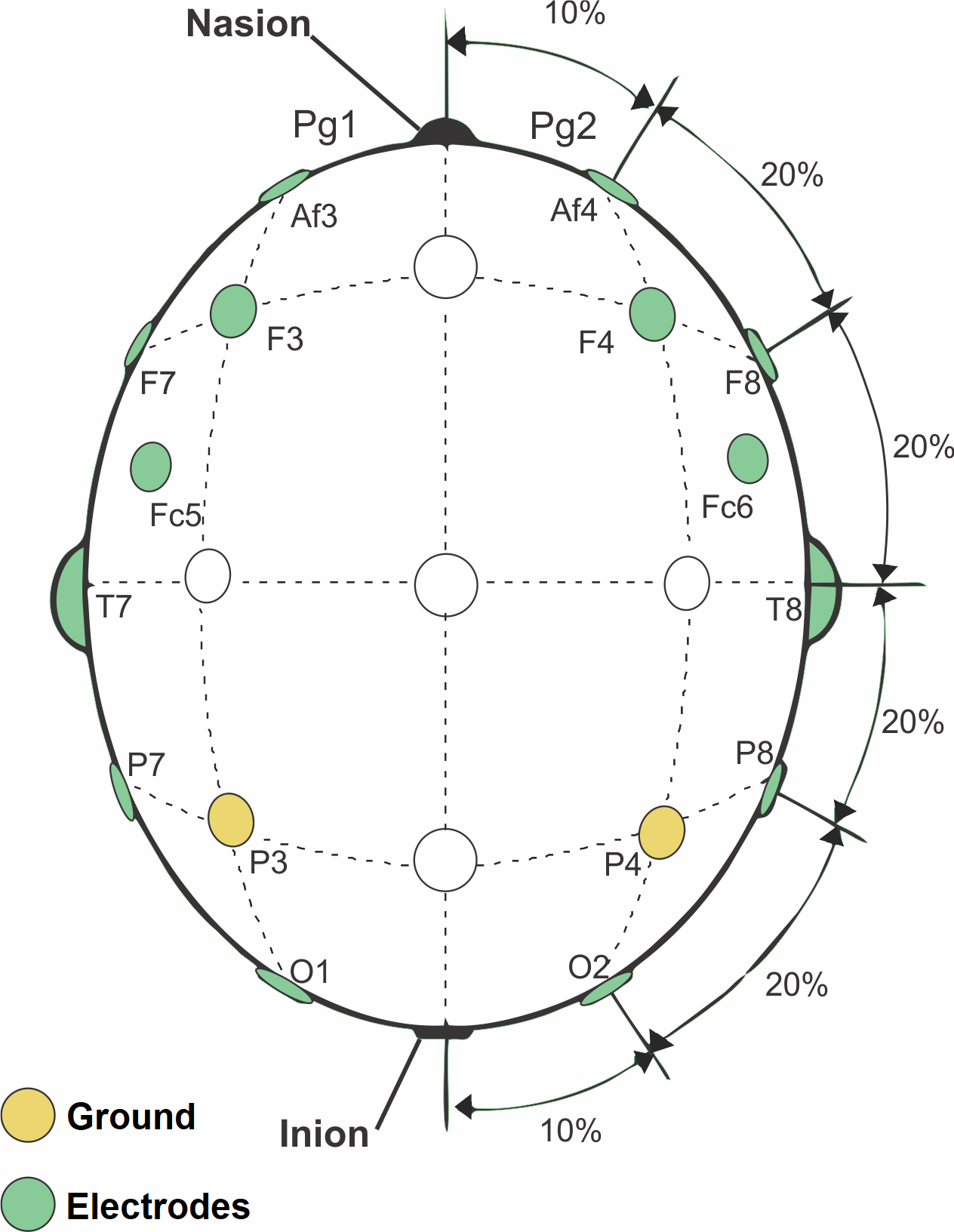}
		\caption{The international 10-20 system to place electrodes} 
		\label{Fig1}
	\end{center}
\end{figure}

The raw data is provided in \cite{AlGoTu:20}. The data were produced by 28 healthy individuals (experimental group) from 20 to 27 years old. An individual plays a game for 5 minutes (20 minutes in total) using the \textit{Emotiv EPOC+} EEG device and earphones. The male-female (M-F) order of players is: F M M M F F M M M M M F M M M M F F F F F M M M M M M M. A single user-independent eGFC model is evolved for the experimental group aiming at reducing individual uncertainty, and enhancing the reliability and generalizability of the system. Brain activity is recorded from 14 electrodes placed on the scalp according to the 10-20 System, namely, at Af3, Af4, F3, F4, F7, F8, Fc5, Fc6, T7, T8, P7, P8, O1, and O2 (Figure \ref{Fig1}). A letter identifies the lobe. F stands for Frontal, T for Temporal, P for Parietal, and O for Occipital. Even and odd numbers refer to positions on the right and left brain hemispheres. The sampling frequency is 128Hz. Each individual produces 38,400 samples per game, and 153,600 samples in total.

\vspace{-2pt}

\vspace{-5pt}

\begin{figure}[h]
	\centering
	\includegraphics[width=8.2cm]{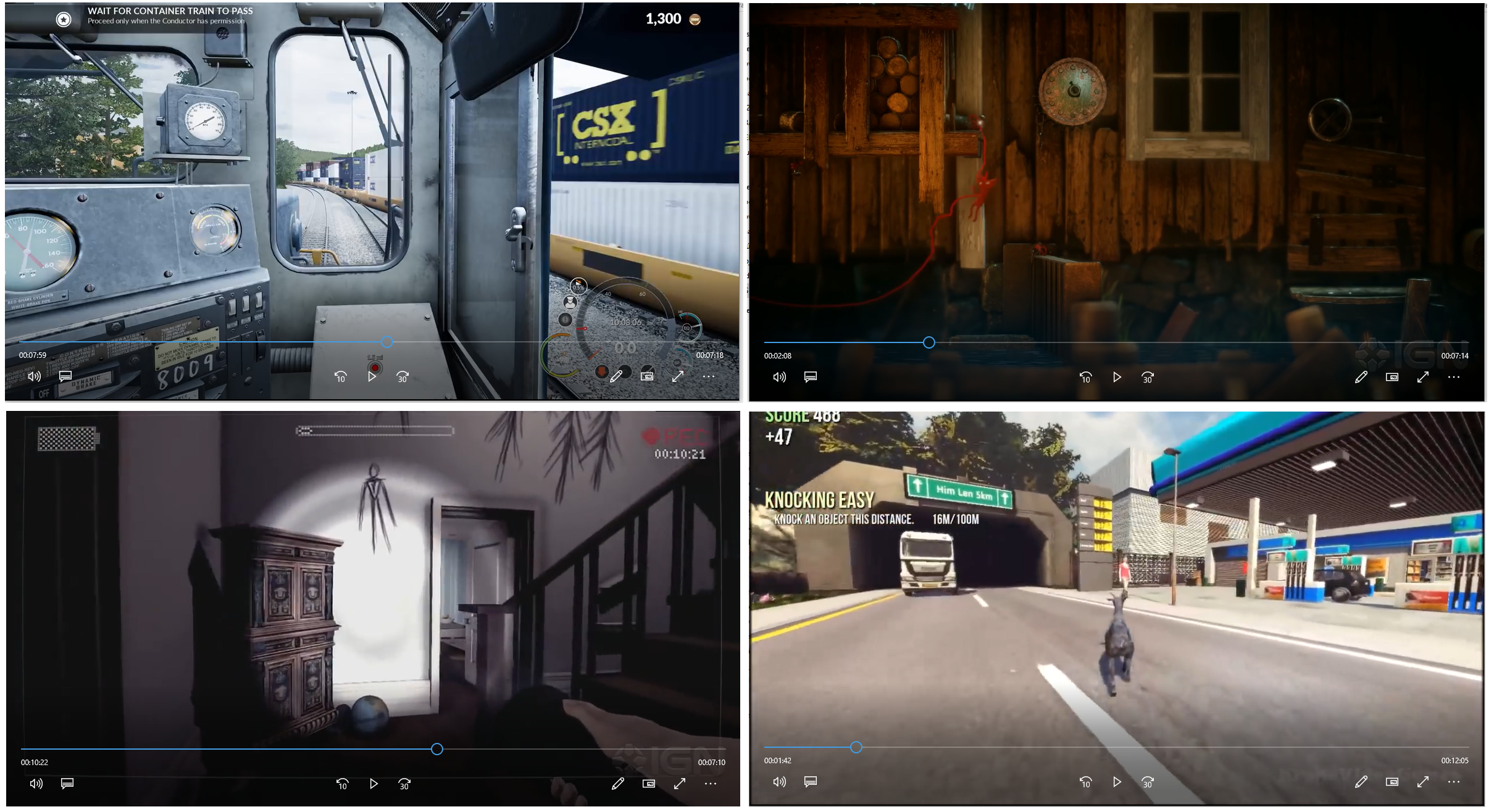}
	\caption{Interfaces of the games Train Sim World, Unravel, Slender The Arrival, and Goat Simulator, used to generate the EEG data streams}
	\label{Fig3}
\end{figure}

The experiments are conducted in a dark and quiet room. A laptop with a 15.6 inch screen, with 16GB high-quality graphic rendering, is used. The games are played in the same order: `Train Sim World', `Unravel', `Slender The Arrival', and `Goat Simulator'. Figure \ref{Fig3} illustrates their interfaces. According to a survey, the predominant class of emotion for the games are, respectively, boredom (Class 1), calmness (Class 2), angriness/nervousness (Class 3), and happiness (Class 4).

\subsection{About Feature Extraction and Experiments}

\begin{figure}[h]
 	\centering
 	\includegraphics[width=8cm]{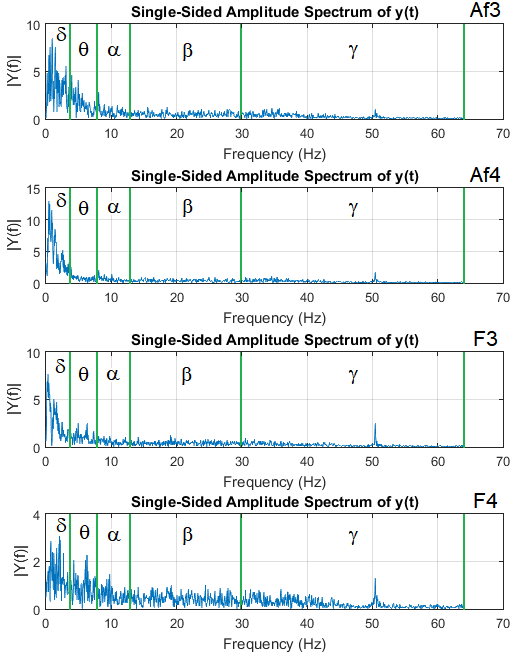}
 	\caption{Examples of spectra and bands obtained from data streams generated by frontal EEG electrodes considering a time window of 30 seconds}
 	\label{Fig4}
\end{figure}

A fifth-order sinc filter is applied to the raw data to suppress movement artifacts \cite{AlGoTu:20}. Subsequently, feature extraction is performed. We generate 10 features from each of the 14 EEG channels. They are the maximum and mean values of 5 bands of the Fourier spectrum. The bands are known as Delta (1-4Hz), Theta (4-8Hz), Alpha (8-13Hz), Beta (13-30Hz), and Gamma (30-64Hz). In case 5-minute time windows are used to construct the frequency spectrum, and compose a processed sample to be fed to eGFC, then each individual produces 4 samples -- one sample per game, i.e., one sample per class or predominant emotion. Therefore, the 28 participants generate 112 processed samples. We also evaluate 1-minute, 30-second, and 10-second time windows, which yield 560, 1120, and 3360 processed samples. Examples of spectra using 30-second time windows over data streams from frontal electrodes -- Af3, Af4, F3 and F4 -- are illustrated in Figure \ref{Fig4}. Notice a higher level of energy in the Delta, Theta and Alpha bands, and that the maximum and mean values per band can be easily obtained to compose a 10-feature processed sample.

In a first experiment, the data of individual electrodes are analyzed (univariate time series analysis). 10-feature samples, $\textbf{x}^{[h]} = [x_1 ~ ... ~ x_{10}]$, $h = 1,...$, are fed to the eGFC, which evolves from scratch. We perform the train-after-test approach, i.e., an eGFC estimate is given; then the true class $C = \{1,2,3,4\}$ of $\textbf{x}^{[h]}$ becomes available, and the pair $(\textbf{x},y)$ is used for a supervised learning step. Classification accuracy, $Acc \in [0,1]$, is obtained recursively from

\begin{equation}
Acc (\textrm{new}) = \frac{h-1}{h} ~ Acc (\textrm{old}) + \frac{1}{h} ~ \tau,
\end{equation}

\noindent in which $\tau := 1$ if the estimate is correct, i.e., $\hat{C}^{[h]} = C^{[h]}$; and $\tau := 0$, otherwise. A random (coin flipping) classifier has an expected $Acc$ of 0.25 (25\%) in the 4-class problem. Higher values indicate a level of learning from the data. From the investigation of individual electrodes we expect to identify more discriminative areas of the brain.

A measure of model compactness is the average number of fuzzy granules over time,

\begin{equation}
c_{avg}(\textrm{new}) = \frac{h-1}{h} ~ c_{avg}(\textrm{old}) + \frac{1}{h} ~ c^{[h]}.
\end{equation}

Second, we consider the global 140-feature problem (multivariate time series analysis) -- being 10 features extracted from each electrode. Thus, $\textbf{x}^{[h]} = [x_1 ~ ... ~ x_{140}]$, $h = 1,...$, are considered as eGFC inputs. The classifier self-develops on the fly based on the train-after-test strategy. Dimension reduction is performed using the non-parametric Spearman's Correlation-based Score method \cite{Soares}. Fundamentally, a feature is better ranked if it is less correlated to other features, and more correlated to a class. The Leave $n$-Feature Out method ($n=5$) gradually eliminates lower ranked features.

\section{Results}

\subsection{Single Channel Experiment}

We evaluate individual EEG channels to discover more propitious brain regions in terms of their relations to emotion patterns. Default hyper-parameters are used to initialize eGFC, i.e. $\rho^{[0]} = \Delta = 0.1$, $h_r = 200$ \cite{FUZZlog2}. Input samples contain 10 entries. Table \ref{tab:ind} shows the accuracy and compactness of eGFC models for different time window lengths. 

\begin{table}[!h] 
	\centering
	\small
	\caption{eGFC classification results for individual electrodes}
	\begin{tabular}{ccc|ccc}
		\hline
		\multicolumn{6}{c}{\textbf{5-minute time window}} \\
		\hline
		\multicolumn{3}{c}{Left hemisphere} & \multicolumn{3}{|c}{Right hemisphere} \\
		\hline
		Ch & $Acc(\%)$ & $c_{avg}$ & Ch & $Acc(\%)$ & $c_{avg}$ \\
		\hline
		Af3     & 18.8  & 24.8  & Af4   & 19.6  & 22.9  \\
		F3      & 23.2  & 21.3  & F4    & 20.5  & 21.2  \\
		F7      & 17.0  & 21.2  & F8    & 25.0  & 19.1  \\
		Fc5     & 24.1  & 22.5  & Fc6   & 20.5  & 19.0  \\
		T7      & 22.3  & 20.2  & T8    & 24.1  & 21.2  \\
		P7      & 18.8  & 21.7  & P8    & 18.8  & 21.7  \\
		O1      & 21.4  & 21.9  & O2    & 21.4  & 21.9  \\
		\hline
		\textbf{Avg.} & 20.8 & 21.9 & \textbf{Avg.} & 21.4 & 21.0 \\
		\hline
		\multicolumn{6}{c}{\textbf{1-minute time window}} \\
		\hline
		\multicolumn{3}{c}{Left hemisphere} & \multicolumn{3}{|c}{Right hemisphere} \\
		\hline
		Ch & $Acc(\%)$ & $c_{avg}$ & Ch & $Acc(\%)$ & $c_{avg}$ \\
		\hline
		Af3     & 43.4  & 17.3  & Af4   & 41.6  & 13.1    \\
		F3      & 37.0  & 14.3  & F4    & 39.6  & 15.5    \\
		F7      & 41.3  & 14.5  & F8    & 31.6  & 10.3    \\
		Fc5     & 38.4  & 16.1  & Fc6   & 41.8  & 18.6    \\
		T7      & 43.9  & 14.4  & T8    & 50.4  & 15.9    \\
		P7      & 37.9  & 17.4  & P8    & 33.8  & 15.7    \\
		O1      & 45.0  & 16.3  & O2    & 40.5  & 16.6    \\
		\hline
		\textbf{Avg.} & 41.0 & 15.8 & \textbf{Avg.} & 39.9 & 15.1 \\
		\hline
		\multicolumn{6}{c}{\textbf{30-second time window}} \\
		\hline
		\multicolumn{3}{c}{Left hemisphere} & \multicolumn{3}{|c}{Right hemisphere} \\
		\hline
		Ch & $Acc(\%)$ & $c_{avg}$ & Ch & $Acc(\%)$ & $c_{avg}$ \\
		\hline
		Af3     & 51.3  & 11.7  & Af4   & 43.9  & 8.9     \\
		F3      & 40.5  & 10.3  & F4    & 42.9  & 8.8     \\
		F7      & 44.2  & 10.3  & F8    & 37.5  & 6.7     \\
		Fc5     & 41.3  & 10.8  & Fc6   & 44.7  & 11.6    \\
		T7      & 40.4  & 10.2  & T8    & 49.6  & 11.3    \\
		P7      & 46.3  &  9.4  & P8    & 42.7  & 10.3    \\
		O1      & 45.9  & 10.1  & O2    & 44.7  & 12.8    \\
		\hline
		\textbf{Avg.} & 44.3 & 10.4 & \textbf{Avg.} & 43.7 & 10.1 \\
		\hline
		\multicolumn{6}{c}{\textbf{10-second time window}} \\
		\hline
		\multicolumn{3}{c}{Left hemisphere} & \multicolumn{3}{|c}{Right hemisphere} \\
		\hline
		Ch & $Acc(\%)$ & $c_{avg}$ & Ch & $Acc(\%)$ & $c_{avg}$ \\
		\hline
		Af3     & 52.6  & 6.3   & Af4   & 44.0  & 5.5   \\
		F3      & 41.8  & 6.1   & F4    & 43.8  & 5.9    \\
		F7      & 46.3  & 6.1   & F8    & 40.2  & 4.8    \\
		Fc5     & 40.5  & 5.7   & Fc6   & 46.8  & 6.8    \\
		T7      & 40.4  & 5.7   & T8    & 51.5  & 6.2    \\
		P7      & 46.9  & 6.7   & P8    & 45.2  & 6.5    \\
		O1      & 47.7  & 6.6   & O2    & 45.2  & 7.0    \\
		\hline
		\textbf{Avg.} & 45.2 & 6.2 & \textbf{Avg.} & 45.2 & 6.1 \\
		\hline
	\end{tabular}
	\label{tab:ind}
\end{table}

From Table \ref{tab:ind} we notice that the mean accuracy for 5-minute windows, $21.1\%$, does not reflect learning. This suggests that the filter effect on extracting features from 5-minute windows suppresses crucial details to distinguish patterns. Emotions tend to be sustained along shorter periods. Accuracies greater than $25\%$ arise for smaller windows. The average accuracy for 1-minute windows, $40.5\%$, is significant, especially due to the limitation of observing a single electrode. As window length reduces, accuracy increases using a more compact model. This is a result of the availability of a larger amount of processed samples (extracted from a smaller amount of time windows), and the ability of the learning algorithm to lead the eGFC to a more stable set up after merging and deleting granules. The difference between the average accuracy of the 30-second ($44.0\%$) and 10-second ($45.2\%$) windows becomes small, which suggests saturation around 45-46\%. Windows smaller than 10 seconds tend to be needless. Asymmetries between the accuracy of models evolved for the left and right brain hemispheres can be noticed. Although the right hemisphere is known to deal with emotional interpretation, and be responsible to creativity and intuition; logical interpretation -- typical of the left hemisphere -- is also present as players seek reasons to justify decisions. Thus, we notice the emergence of patterns in both hemispheres.

In general, with focus on the 30- and 10-second windows, the pair of frontal electrodes, Af3-Af4, gives the best recognition results, especially Af3. The frontal cortex includes the premotor and primary motor cortices, which controls voluntary movements of body parts. Thus, patterns that arise from the Af3-Af4 streams may be strongly related to brain commands to move hands and arms -- which is indirectly related to the emotions of the players. A spectator, instead of a game actor, may have a classification model based solely on Af3-Af4 with diminished accuracy. The pair Af3-Af4 is closely followed by the pairs O1-O2 (occipital lobe) and T7-T8 (temporal lobe), which is somehow consistent with the results in \cite{Xiang:18} \cite{Zheng}. The temporal lobe is useful for audition, language processing, declarative memory, and visual and emotional perception. The occipital lobe contains the primary visual cortex and areas of visual association. Neighbor electrodes, P7-P8, also offer relevant information for classification.

\subsection{Multiple Channel Experiment}

\begin{figure}[!h]
	\centering
	\includegraphics[width=8cm]{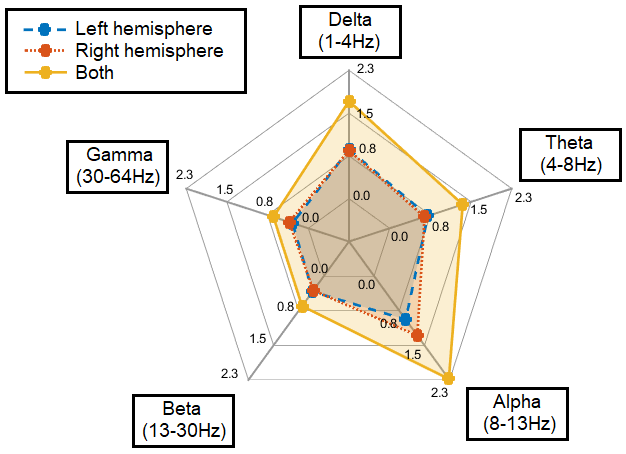}
	\caption{Spearman correlation between frequency bands and emotion classes}
	\label{Fig5}
\end{figure}

The multivariate 140-feature stream is fed to eGFC. We consider 10-second windows over 20-minute recordings per individual -- the best set up secured in the previous experiment. Features are ranked from the Spearman's Correlation Score. To give quantitative evidence, we sum the monotonic correlations between a band and the classes of emotions (a sum of 14 items that correspond to the 14 EEG channels). Figure \ref{Fig5} shows the result in dark yellow, and its decomposition per brain hemisphere. The global values are precisely: 2.2779, 1.7478, 1.3759, 0.6639, and 0.6564 for the Alpha, Delta, Theta, Beta, and Gamma bands, respectively. The higher, the better.

We apply the strategy of leaving $5$ features out to evaluate the user-independent eGFC. Each eGFC model delineates class boundaries in a different multi-dimensional space. Table \ref{tab:Spearman} shows the results for the multi-channel streams. A quadcore laptop, i7-8550U, with 1.80GHz, 8GB RAM, and Windows 10 OS, is used. We testify that  brain activity patterns are found in all frequency bands, viz., Delta, Theta, Alpha, Beta, and Gamma, as even lower ranked features positively affect the estimates. Features from any channel may assist the classifier decision. The greatest eGFC accuracy, $72.20\%$, occurs using all features extracted from the spectrum. In this case, a stream of 3,360 samples is processed in 6.032 seconds (1.8ms per sample). As a sample is generated at every 10s (10-second windows), and processed in 1.8ms, we have shown that eGFC can operate in real time with much larger rule bases and Big data, such as data generated by additional electrodes, and other physiological and non-physiological means.

\vspace{-3pt}

\begin{table}[ht]
	\centering
	\normalsize
	\caption{eGFC classification results in online emotion recognition}
	\begin{tabular}{c|ccc}
		\hline
		\# Features & $Acc~(\%)$ & $c_{avg}$ & CPU Time (s) \\ 
		\hline
		140 & 72.20 & 5.52 & 6.032 \\
		135 & 71.10 & 5.70 & 5.800 \\
		130 & 71.46 & 5.68 & 5.642 \\
		125 & 71.25 & 5.75 & 5.624 \\
		120 & 70.89 & 5.79 & 5.401 \\
		115 & 70.92 & 5.68 & 5.193 \\
		110 & 70.68 & 5.73 & 4.883 \\
		105 & 70.68 & 5.71 & 4.676 \\
		100 & 70.24 & 5.78 & 4.366 \\
		95 & 70.12 & 5.69 & 4.290 \\
		90 & 69.91 & 5.66 & 3.983 \\
		85 & 66.31 & 5.53 & 3.673 \\
		80 & 65.09 & 5.45 & 3.534 \\
		75 & 65.48 & 5.26 & 3.348 \\
		70 & 64.08 & 5.26 & 2.950 \\
		65 & 63.42 & 5.37 & 2.775 \\
		60 & 62.56 & 5.33 & 2.684 \\
		55 & 60.95 & 5.33 & 2.437 \\
		50 & 60.57 & 5.57 & 2.253 \\
		45 & 60.39 & 5.53 & 2.135 \\
		40 & 54.13 & 5.14 & 1.841 \\
		35 & 52.29 & 5.28 & 1.591 \\
		30 & 52.52 & 5.54 & 1.511 \\
		25 & 50.95 & 5.61 & 1.380 \\
		20 & 51.42 & 5.58 & 1.093 \\
		15 & 49.28 & 5.57 & 0.923 \\
		10 & 48.63 & 5.58 & 0.739 \\ 
		\hline
	\end{tabular}
	\label{tab:Spearman}
\end{table}

\begin{figure}[t]
	\centering
	\includegraphics[width=8cm]{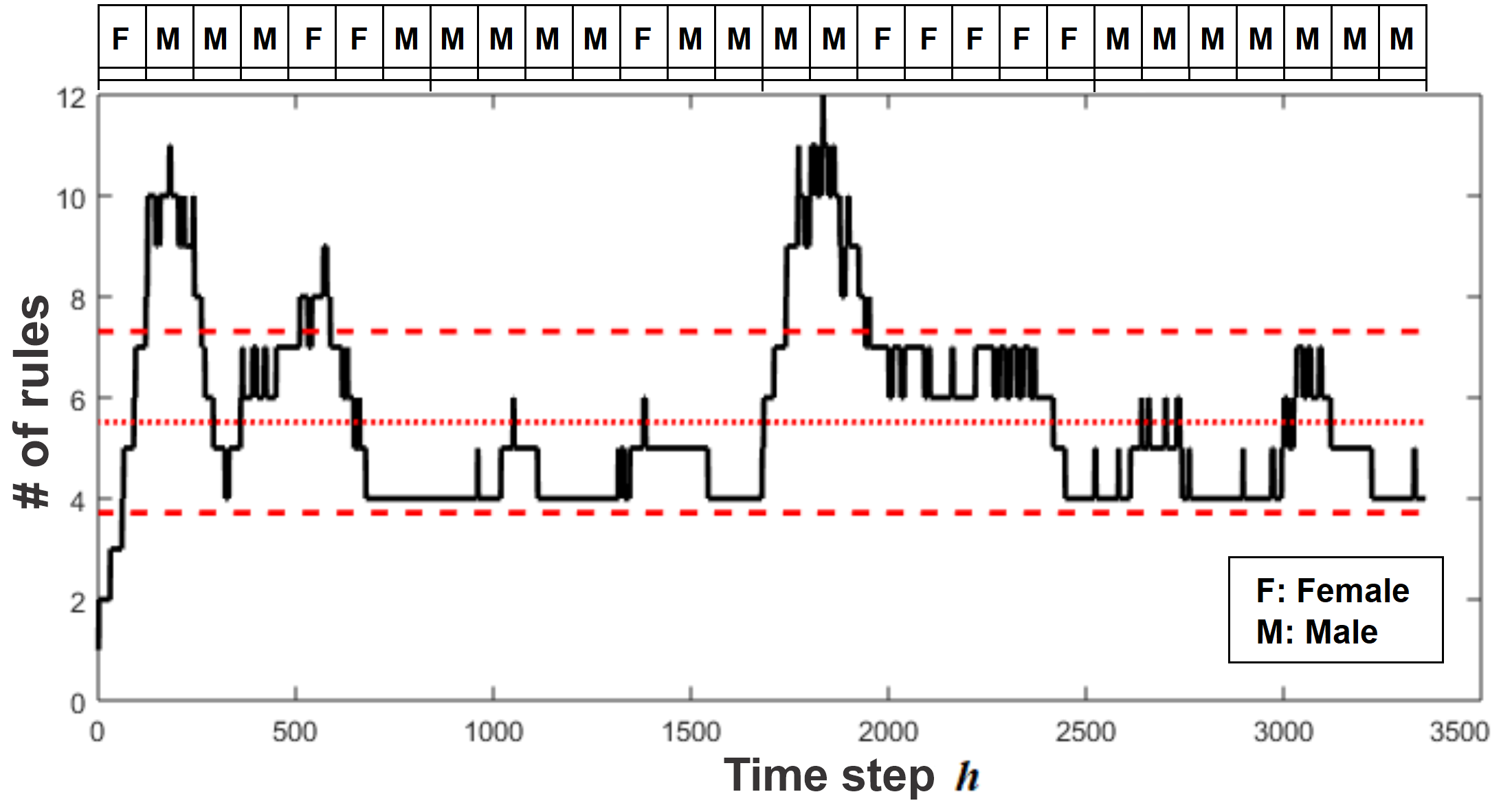}
	\vspace{-2pt}
	\caption{Structural evolution of the best user-independent eGFC model}
	\label{Fig6}
\end{figure}

The average number of fuzzy rules along the learning steps, around 5.5, is similar for all streams analyzed (Table \ref{tab:Spearman}). An example of eGFC structural evolution (best case, $72.20\%$) is shown in Figure \ref{Fig6}. Notice that the emphasis of the model and algorithm is to keep up with accuracy during online operation at the price of merging and deleting rules occasionally. If a higher level of memory is desired, then the default eGFC hyper-parameters can be changed by turning off the rule removing threshold, $h_r = \infty$; and setting the merging parameter, $\Delta$, to a lower value. Figure \ref{Fig6} also shows that the effect of male-female shifts due to the consecutive use of the EEG device does not imply a need of completely new granules and rules. Parametric adaptation of Gaussian granules is often enough to accommodate slightly different behaviors. Nevertheless, if a higher level of memory is allowed, rules for both genders can be kept separately. In general, the results using evolving fuzzy intelligence and EEG streams as unique source of physiological data are encouraging.

\section{Conclusion}

We have evaluated online learning and an evolving fuzzy classifier, called eGFC, for recognition of emotions in games. The players are subject to visual and auditory stimuli using an EEG device. An emotion (boredom, calmness, angriness, or happiness) prevails in each game, according to the Arousal-Valence system. 140 features are extracted from the frequency spectra of 14 electrodes/brain regions. We analyzed the Delta, Theta, Alpha, Beta, and Gamma bands. We examined the contribution of single electrodes on emotion recognition, and the effect of window lengths and dimensionality reduction on the overall accuracy of the eGFC models. We conclude: (i) electrodes on both brain hemispheres assist the recognition of emotions expressed through a variety of spatio-temporal patterns; (ii) the frontal (Af3-Af4) area, followed by the occipital (O1-O2) and temporal (T7-T8) areas are, in this order, slightly more discerning than the others; (iii) although patterns may be found in any band, the Alpha (8-13Hz) band, followed by the Delta (1-4Hz) and Theta (4-8Hz) bands, is more monotonically correlated with the emotion classes; (iv) eGFC is able to learn from and process 140-feature samples in 1.8 milliseconds per sample. Thus, eGFC is suitable for real-time applications considering EEG and other data sources; (v) a greater number of features covering the entire spectrum and the use of 10-second windows guided eGFC to its best accuracy, 72.20\%, using from 4 to 12 granules and rules. In the future, wavelet transforms will be evaluated. A deep neural network (as feature extractor) will be connected to eGFC for real-time emotion recognition. We also intend to evaluate an ensemble of customized evolving classifiers.

\section*{Acknowledgement}

This work received support from the Serrapilheira Institute (Serra - 1812-26777).

\bibliographystyle{IEEEtran}

\bibliography{Biblio}

\end{document}